\documentclass[lettersize,journal]{IEEEtran}
\usepackage{amsmath,amsfonts}
\usepackage{algorithmic}
\usepackage{array}
\usepackage[caption=false,font=normalsize,labelfont=sf,textfont=sf]{subfig}
\usepackage{textcomp}
\usepackage{stfloats}
\usepackage{url}
\usepackage{verbatim}
\usepackage{graphicx}
\def\BibTeX{{\rm B\kern-.05em{\sc i\kern-.025em b}\kern-.08em
    T\kern-.1667em\lower.7ex\hbox{E}\kern-.125emX}}
\usepackage{balance}

\usepackage{multirow} 
\usepackage{booktabs} 
\usepackage{pgfplots}
\usepackage{subcaption}
\usepackage[linesnumbered,ruled,vlined]{algorithm2e}
\usepackage{fancyhdr}

\begin{document}

\title{WECAR: An End-Edge Collaborative Inference and Training Framework for WiFi-Based Continuous Human Activity Recognition}

\author{
Rong Li, Tao Deng, Siwei Feng, He Huang, Juncheng Jia, Di Yuan, and Keqin Li,~\IEEEmembership{Fellow,~IEEE}

\thanks{This paper was partially presented at AAAI Conference on Artificial Intelligence (AAAI) 2025 \cite{rong2025consense}.}
\thanks{R. Li, T.\ Deng, S.\ Feng, H.\ Huang, and J.\ Jia are with School of Computer Science and Technology, Soochow University, Suzhou, Jiangsu 215006, China (E-mail:  20235227092@stu.suda.edu.cn, \{dengtao;~swfeng;~huangh;~jiajuncheng\}@suda.edu.cn.)

D.\ Yuan is with the Department of Information Technology, Uppsala University, 751 05 Uppsala, Sweden (E-mail: di.yuan@it.uu.se)

K.\ Li is with the Department of Computer Science, State University of New York, New Paltz, New York 12561, USA (E-mail:
lik@newpaltz.edu)

}



}

\maketitle

\begin{abstract}
WiFi-based human activity recognition (HAR) holds significant promise for ubiquitous sensing in smart environments. 
A critical challenge lies in enabling systems to dynamically adapt to evolving scenarios, learning new activities without catastrophic forgetting of prior knowledge, while adhering to the stringent computational constraints of edge devices. 
Current approaches struggle to reconcile these requirements due to prohibitive storage demands for retaining historical data and inefficient parameter utilization.
We propose WECAR, an end-edge collaborative inference and training framework for WiFi-based continuous HAR, which decouples computational workloads to overcome these limitations.
In this framework, edge devices handle model training, lightweight optimization, and updates, while end devices perform efficient inference.
WECAR introduces two key innovations, i.e., dynamic continual learning with parameter efficiency and 
hierarchical distillation for end deployment.
For the former, we propose a transformer-based architecture enhanced by task-specific dynamic model expansion and stability-aware selective retraining.
For the latter, we propose a dual-phase distillation mechanism that includes multi-head self-attention relation distillation and prefix relation distillation.
We implement WECAR based on heterogeneous hardware using \textit{Jetson Nano} as edge devices and the \textit{ESP32} as end devices, respectively.
Our experiments across three public WiFi datasets reveal that WECAR not only outperforms several state-of-the-art methods in performance and parameter efficiency, but also achieves a substantial reduction in the model's parameter count post-optimization without sacrificing accuracy.
This validates its practicality for resource-constrained environments.
\end{abstract}


\section{Introduction}
\begin{figure}[!t]
\centering
\includegraphics[width=0.7\columnwidth]{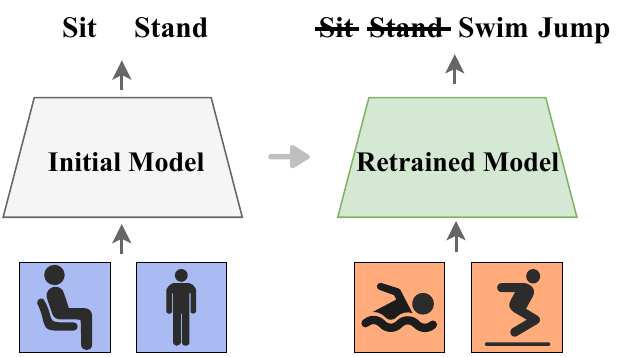}
\caption{The initial model was trained to recognize two activities: sitting and standing. 
However, after retraining the model with data from two new activities swimming and jumping, it loses the ability to recognize sitting and standing.}
\label{fig_issue1}
\vspace{-0.45cm} 
\end{figure}

Human activity recognition (HAR) technologies are widely used in medical monitoring \cite{6477043}, smart homes \cite{9097296}, and security \cite{10842509}.
Traditional video surveillance systems for HAR face challenges like privacy issues, limited field of view, and lighting sensitivity. 
In contrast, wireless signal sensing, particularly WiFi-based HAR, offers a non-intrusive and privacy-respecting solution. 
WiFi is ideal for HAR due to its ubiquity and low hardware requirements \cite{9055083, 10138672}. 
At present, WiFi-based HAR mainly relies on channel state information (CSI) that describes how the signal propagates from the WiFi transmitter to the receiver, including attenuation, scattering, fading, and power changes with distance on each WiFi transmission path. 
For WiFi-based HAR, traditional deep learning (DL) models \cite{8884151, 8514811} face difficulties in recognizing new activities, as they rely on static models that cannot adapt to new human activities.
Simple fine-tuning of DL models with new data without access to the original training data may result in catastrophic forgetting, where previously learned knowledge is overwritten.
As shown in Fig. \ref{fig_issue1}, the model retrained on new activities (swim and jump) fails to recognize prior activities (sit
and stand).
Therefore, there is an urgent need to develop DL models that can effectively adapt to dynamically changing environments. 

In practical WiFi deployments, WiFi receivers serve as the primary data acquisition units, capturing detailed CSI from ambient signals. 
Due to cost considerations and practical constraints, these devices are typically resource-constrained (e.g., the ESP32-C3, which can be used directly as a WiFi receiver, is equipped with a RISC-V 32-bit single-core processor, 400 KB SRAM and 4 MB flash memory \cite{cameron2023esp32}), limiting their capabilities to model inference while rendering model training infeasible. This constraint inherently establishes a dual-layer architecture that decouples training from inference \cite{10623307}.
In addition, these constraints necessitate WiFi-based HAR systems to be capable of continuous sensing without storing user data, which is a critical requirement for preserving privacy and operational efficiency.
In order to address privacy and storage issues,
exemplar-free class-incremental learning (EFCIL) \cite{li2017learning,zhu2022self,goswami2024resurrecting} aims to recognize both old and new activity classes without retaining previous exemplars.
However, while existing EFCIL methods have demonstrated their efficacy in the area of computer vision (CV), their direct adaptation to WiFi-based HAR introduces unique challenges. 
First, unlike static images, WiFi signals experience subtle, time-sensitive fluctuations due to human activities, which complicates the extraction of stable features. The continuous and rapidly changing nature of WiFi data, coupled with the lack of clear spatial references, further exacerbates this issue. Additionally, conventional EFCIL approaches typically demand significant computational resources and extended training durations, rendering them impractical for deployment on resource-constrained edge devices like WiFi receivers.
Consequently, there exists an urgent need for a lightweight framework that simultaneously meet three critical requirements: (1) effective spatiotemporal feature extraction from sequential WiFi data, (2) mitigation of catastrophic forgetting in incremental learning scenarios, and (3) substantial reduction of computational and storage overhead for edge compatibility.

In this work, we propose an end-edge collaborative inference and training EFCIL framework for WiFi-Based HAR. We name the framework WECAR, as it involves WiFi, end-edge devices, continual learning, and activity recognition.
In WECAR, model training, lightweighting, and updates occur on edge devices, while inference is delegated to end devices.
The main contributions of this paper are as follows.
\begin{enumerate}
    \item
To capture spatiotemporal dependencies in sequential WiFi data, we leverage a transformer-based architecture augmented with two synergistic mechanisms, i.e., dynamic model expansion and selective retraining.
In the dynamic model expansion mechanism, we introduce parameter-efficient task-specific prefixes – small-scale trainable parameters embedded within multi-head self-attention (MHSA) layers. 
These learnable prefixes are optimized per incremental task to isolate and preserve critical task-specific features while preventing interference with previously acquired knowledge.
In the selective retraining mechanism, a neuron stability-aware weighting mechanism dynamically adjusts multilayer perceptron (MLP) layer contributions based on cross-task performance metrics.
By prioritizing stable neurons for retention and adaptive ones for recalibration, this approach balances plasticity for new tasks with stability for past knowledge.
This dual strategy enables rapid training through partial parameter updates while maintaining robust continual learning performance.
    \item
To enable efficient edge-to-end device deployment, we develop a two-tier distillation framework, i.e., MHSA relation distillation and prefix relation distillation.
The MHSA relation distillation aligns the student model’s attention patterns with the teacher model by minimizing discrepancies in attention matrices across all heads, preserving inter-feature relationships critical for temporal dynamics.
The prefix relation distillation transfers task-specific prefix embeddings from the teacher to student models through contrastive alignment in the MHSA latent space, ensuring retention of historical task knowledge during model compression.
This paradigm reduces model parameters without significant accuracy degradation, producing deployable models that retain both generalization capacity and incremental learning readiness.
    \item
We implement WECAR using \textit{Jetson Nano} and \textit{ESP32} as edge and end devices respectively.
We validate WECAR on three publicly available WiFi datasets.
The results show that it outperforms state-of-the-art EFCIL baselines in average incremental accuracy while requiring fewer parameters.
In addition, it maintains baseline accuracy despite parameter reduction, validating its practical viability for resource-constrained environments.


\end{enumerate}

\section{Related Work}
The field of WiFi-based HAR has gained extensive attention. 
For instance, the works in \cite{gu2014wifi} and \cite{Abdelnasser2015} use the received signal strength indication (RSSI) of WiFi for gesture and activity recognition. 
However, RSSI’s effectiveness is constrained by multipath and fading effects. 
In contrast, WiFi's CSI can provide more reliable information, as highlighted in \cite{10480395}.
This has led to its adoption in works such as \cite{Zhang2017}, which employs the Fresnel zone model to enhance recognition accuracy.
To bypass the limitations of manual feature extraction, DL based methods have been increasingly utilized. 
For example, \cite{Wang2019Can} applies 1D Convolutional Neural Networks (CNNs) for indoor positioning, while \cite{Chahoushi2023} uses CNNs to classify activities from CSI. Despite these advancements, traditional DL models are static models that cannot adapt to new human actions.


Class-incremental learning (CIL) emerges as a promising solution to address the dynamic adaptation challenges in HAR.
 CIL can be broadly categorized into three types:
 rehearsal-based methods \cite{rebuffi2017icarl, wu2019large}, regularization-based methods \cite{kirkpatrick2017overcoming, yang2021cost}, and dynamic architecture-based methods \cite{rusu2016progressive, douillard2022dytox}. 
These approaches primarily rely on storing exemplars from past tasks to mitigate catastrophic forgetting. 
However, in real-world WiFi systems, receivers are often resource-constrained and face privacy concerns in user data retention. 
This makes data storage impractical and calls for WiFi-based HAR systems that can operate continuously without storing user data, ensuring both privacy and efficiency.
To tackle this issue, several EFCIL methods have been proposed.
For example, the work in \cite{li2017learning} integrates knowledge distillation into CIL, although its efficacy is limited when relying solely on new data. 
The work in \cite{gao2022r} proposes a framework that separates representation and classifier learning, thereby improving data synthesis for previous tasks.
The work in \cite{asadi2023prototype} introduces a prototype-sample relation distillation by combining supervised contrastive loss \cite{khosla2020supervised}, self-supervised learning \cite{liu2021self}, and class prototype evolution techniques \cite{de2021continual}. 
These methods mainly employ ResNet and other CNN-based models.
However, CSI data's dimensions (time stamps and channel state) are fundamentally different from the spatial dimensions of image data. 
Consequently, using a 2D kernel to extract spatial patterns from CSI data results in suboptimal feature extraction. While numerous CNN and ResNet-based approaches have been developed for EFCIL, transformer-based EFCIL remains a relatively unexplored area. 
The work in \cite{roy2023exemplar} adapts transformer MHSA layers with convolution for new tasks, but it is unsuitable for WiFi data due to its reliance on image augmentation.

A few recent works investigate WiFi-based HAR with incremental learning \cite{zhang2023csi,ding2023passive,fu2024ccs}. The work in \cite{zhang2023csi} employs exemplars and distillation loss for knowledge retention, though its reliance on stored exemplars raises privacy concerns and memory constraints. The work in \cite{ding2023passive} introduces an enhanced CNN with attention and dual-loss functions. However, \cite{ding2023passive} is limited to processing one category at a time, restricting its applicability. 
The work in \cite{fu2024ccs} proposes a solution tailored for wireless sensing services, suitable for customized incremental services, but it also requires storing exemplars.

We previously proposed a ConSense model for WiFi-based HAR \cite{rong2025consense}. 
Compared to \cite{rong2025consense}, the contributions of this paper are as follows.
First, while \cite{rong2025consense} focuses on a single-device continual learning, this paper introduces a novel end-edge collaborative paradigm that decouples training/lightweighting (edge) from inference (end devices). 
This architectural shift enables privacy preservation, reduces on-device computational burden, and supports scalable deployment; such advanced have not been addressed in prior work.
In addition, to enable efficient edge-to-end device deployment, this paper develops a two-tier distillation framework.
This framework addresses a critical gap in \cite{rong2025consense} by enabling task-aware model compression and zero exemplar leakage during edge-to-end device deployment, thus ensuring strict privacy compliance.
Finally, this paper realizes an end-to-end system.
Unlike the simulation-based validation in \cite{rong2025consense}, this paper demonstrates real-world practicality through implementation on heterogeneous hardware (Jetson Nano edge and ESP32 end devices).


\section{Background and Problem Statement}
\subsection{WiFi Channel State Information}
The principle of WiFi-based HAR stems from the dynamic interplay between propagating radio waves and human motion. 
When transmitted signals interact with environmental perturbations and obstacles caused by human activities, alterations in propagation paths occur through reflection, diffraction, and scattering phenomena. These path deviations induce time-variant multipath interference patterns, encoding motion-induced environmental signatures into measurable signal distortions.
By analyzing these time-variant signal distortions, we can derive precise environmental sensing.

CSI quantifies fine-grained signal propagation characteristics in multipath WiFi channels, capturing physical-layer distortions such as frequency-selective fading, delay spread, and Doppler shifts. 
Leveraging the inherent subcarrier granularity of orthogonal frequency division multiplexing (OFDM), CSI enables multidimensional feature extraction by decomposing channel responses across orthogonal subcarriers. This process is formally modeled as
\begin{equation}
H(f_k) = \lvert H(f_k) \rvert e^{j\angle H(f_k)},
\end{equation}
where $H(f_k)$ denotes the CSI of the subcarrier with central frequency of $f_k$ and $\angle H$ denotes its phase. 
The value of each CSI is represented in complex form as $a + bi$, with its amplitude expressed as $\sqrt{a^2 + b^2}$.
The CSI is organized into a matrix $\mathbf{D} \in \mathbb{R}^{n \times d}$.
Here, $n$ represents the temporal dimension, defined as $n = fs$, where $f$ denotes the frequency of data packet collection, and $s$ represents the sampling time for an action.
The channel dimension $d$ is defined as $d = ew$, where $e$ refers to the number of transmitting and receiving antennas, and $w$ represents the number of subcarriers per antenna pair.

\begin{figure}[!t]
\centering
\includegraphics[width=0.98\columnwidth]{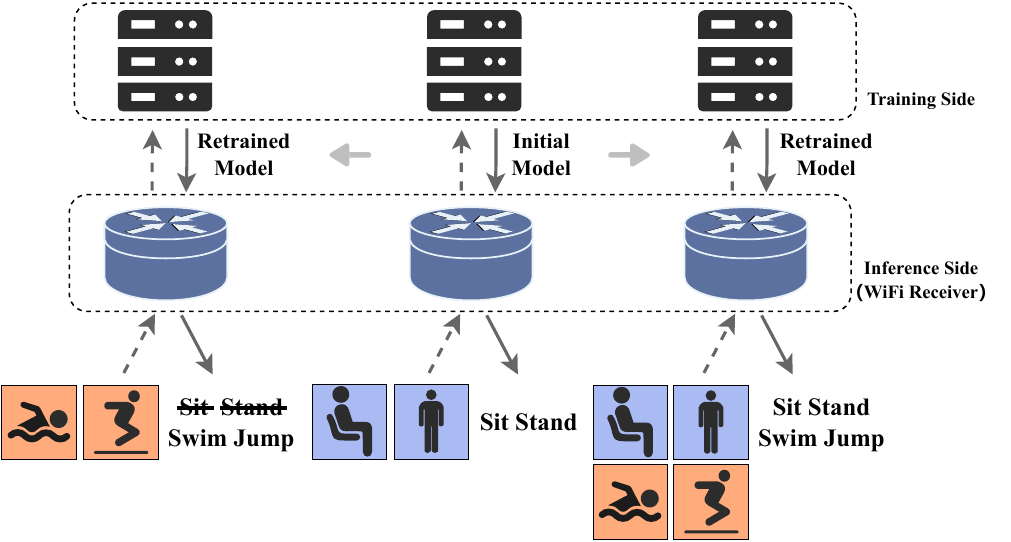}
\caption{The middle layer includes WiFi receivers, which are responsible for data collection and inference. 
The upper layer represents the server, which is in charge of model training. 
After the receiver collects the CSI data, it transmits it to the server for model training. 
Once the training is complete, the trained model is sent back to the receiver for inference.}
\label{fig_issue2}
\vspace{-0.45cm} 
\end{figure}

\subsection{Problem Statement}
In practical WiFi-based sensing systems, the WiFi receiver serves as the principal data acquisition node, capturing discrete activity classes set (denoted by $\mathcal{C}_t$, where $t$ represents the t-th data collection) at each sampling interval.
For example, as shown in the central panel of Fig. \ref{fig_issue2}, the initial data collection phase detects two distinct activities (sitting and standing).
While terminal devices are constrained by limited storage and computational capacity for local model training, cloud-centric approaches, though theoretically capable of addressing computational bottlenecks, introduce prohibitive privacy vulnerabilities and bandwidth overhead due to frequent model updates and bulk data transmission. 
To mitigate these challenges, we delegate the training process to a co-located edge module. 
This architecture ensures privacy preservation through localized data processing and reduces bandwidth demands via lightweight model optimization. 
The optimized models are then deployed on terminals for real-time inference.
Thus, from the perspectives of real-time performance, overhead, and privacy protection, it is preferable to decouple inference and training.


Traditional DL models face inherent limitations in WiFi-based HAR scenarios, particularly in recognizing emerging activities. Static DL models lack adaptability to novel actions, and naive fine-tuning with new data (absent original training data) triggers catastrophic forgetting—a phenomenon where parameter overwriting erases prior knowledge.
For example, when the WiFi receiver detects two additional activities (swimming and jumping, see the left panel of Fig. \ref{fig_issue2}), retraining the DL model disrupts its ability to recognize previously learned classes (sitting and standing).
Conventional mitigation strategies (see the right panel of Fig. \ref{fig_issue2}) necessitate joint retraining with historical and new data to retain prior knowledge, but this approach remains impractical under privacy regulations and device storage constraints.


At present time, the critical problem is: 
\textbf{How can resource-constrained devices retain real-time recognition capabilities for historical activities without storing raw data?}

\begin{figure*}[htbp]
\centering
\includegraphics[width=0.8\textwidth]{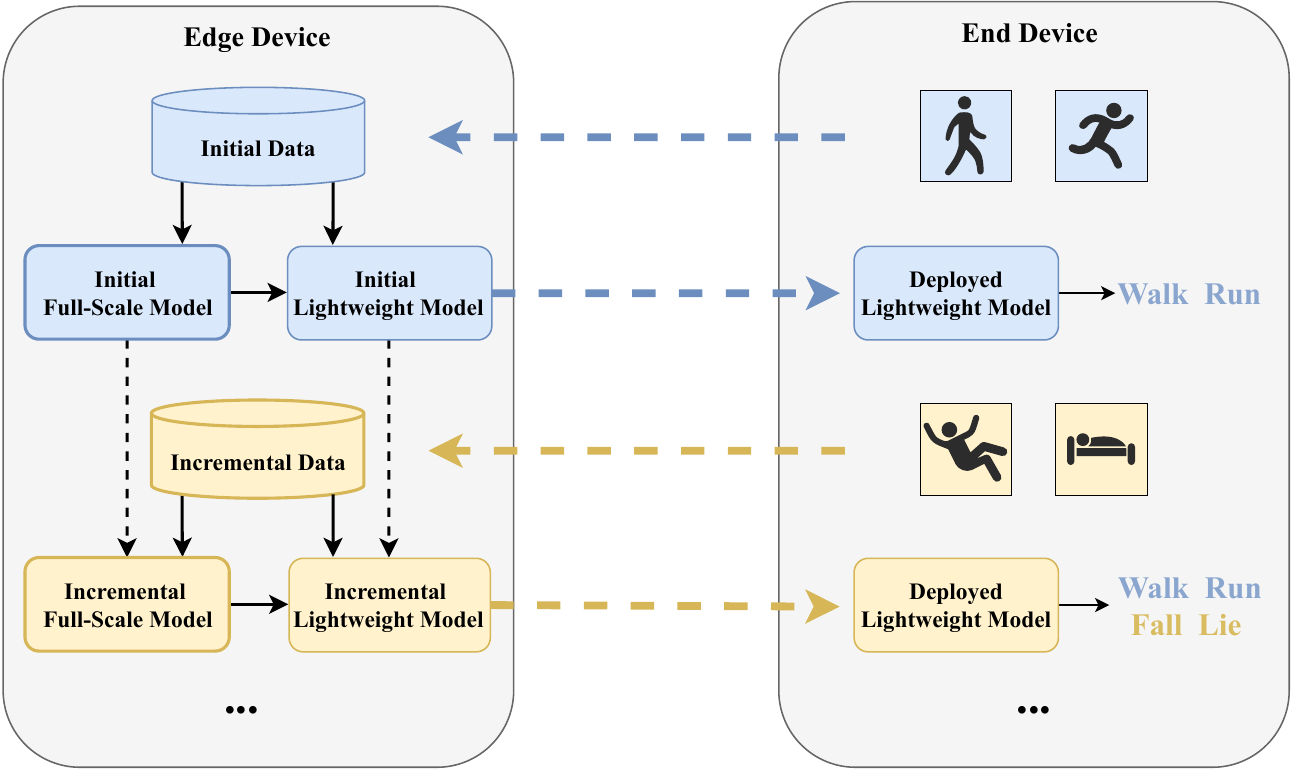} 
\caption{Framework overview.}
\label{fig_model_part1}
\end{figure*}

\section{System Design}

\begin{table}[!htbp]
    \caption{List of Main Notations}
    \centering
    \begin{tabular}{ll}
        \hline
        Notation & Definition \\ \hline
        $D$ & The matrix of CSI  \\
        $n$ & The temporal dimension of CSI  \\
        $d$ & The channel dimension of CSI  \\
        $X$ & The original stream \\
        $X'$ & The stream processed by Gaussian positional encoding \\
        $h$ & The number of attention heads \\
        $Q$ & The query matrix \\
        $K$ & The key matrix \\
        $V$ & The value matrix \\
        $t$ & The task index \\
        $P^t_K$ & The trainable key prefix parameters for task $t$ \\
        $P^t_V$ & The trainable value prefix parameters for task $t$ \\
        $P^{t-1}_{K,frozen}$ & The frozen key prefix parameters from the previous task \\ 
        $P^{t-1}_{V,frozen}$ & The frozen value prefix parameters from the previous task \\ \hline
    \end{tabular}
    \label{tab:parameter}
\end{table}

\subsection{Overall Framework}
In order to solve the above problem, we propose WECAR, an end-edge collaborative inference
and training framework, for WiFi-based continuous HAR, as shown in Fig. \ref{fig_model_part1}.
WECAR establishes a hierarchical architecture where edge devices execute model training, lightweight optimization, and updates, while end devices conduct efficient inference. 
The framework operates through two core models, i.e., the full-scale model (FSM) and the lightweight model (LWM).
In the FSM, the original unoptimized model with complete network architecture and parameters are present.
The LWM has instead a compressed variant derived from FSM through  lightweight.

The interaction process between edge devices and end devices mainly includes four stages, i.e., the initial training stage, initial lightweight stage, incremental training stage, and incremental lightweight stage.
In the initial training stage $T_1$, the edge device trains FSM on the first activity class set $\mathcal{C}_1$, updating all parameters (MHSA layers, MLP, and classifier). 
When the initial training stage is completed, MHSA layer parameters $W^{(MHSA)}_{frozen} = \{W^Q_{frozen}, W^K_{frozen}, W^V_{frozen}\}$ are fixed to preserve learned representations.
In the initial lightweight stage, FSM serves as a teacher to distill knowledge into the student model LWM, which shares FSM's architecture but employs an MLP with reduced dimension.
The MHSA relation distillation leverages three losses, i.e., attention relation distillation loss $\mathcal{L}_{AT}$, value relation distillation loss $\mathcal{L}_{VR}$ and logits distillation loss $\mathcal{L}_{log}$.
Once the initial lightweight stage is completed, LWM's MHSA weights become aligned with FSM's frozen parameters. 
The optimized LWM is deployed to end devices for inference.
In the incremental training stage at session $t$, for new class set $\mathcal{C}_t$, WECAR dynamically adapts FSM through MHSA expansion and MLP selective retraining.
Specifically, MHSA expansion introduce trainable prefixes
$P^t = \{P^t_{K},P^t_{V}\}$ while retaining historical prefixes $P^{t-1}_{frozen}$, where $P^{t-1}_{frozen} = \{P^{t-1}_{K,frozen}, P^{t-1}_{V,frozen}\}$ represent the frozen key and value prefixes from the previous task.
MLP selective retraining freezes stable neurons identified via activation analysis and updates only the unstable neurons.
After completing task $t$, the previous task prefixes $P^{t-1}_{frozen}$ are updated to incorporate the new prefixes from task $t$. Specifically, $P^{t-1}_{frozen}$ is updated as follows: $P^t_{frozen} = \{ P^{t}_{K,frozen}, P^{t}_{V,frozen} \}$, where $P^{t}_{K,frozen} = [P^{t}_K, P^{t-1}_{K,frozen}]$ and $P^{t}_{V,frozen} = [P^{t}_V, P^{t-1}_{V,frozen}]$, where [,] denotes the concatenation operation.
In the incremental lightweight stage, 
in order to reduce the number of parameters, the student model’s MHSA layer does not retain the prefixes from previous tasks.
Knowledge transfers from the teacher model FSM to the student model LWM takes place via prefix relation distillation loss $L_P$ and logits distillation loss $L_{log}$.
The lightweighted incremental model LWM is then processed and transmitted to the end device for inference.
In the following, we will introduce WECAR in more detail.

\begin{figure*}[htbp]
\centering
\includegraphics[width=0.98\textwidth]{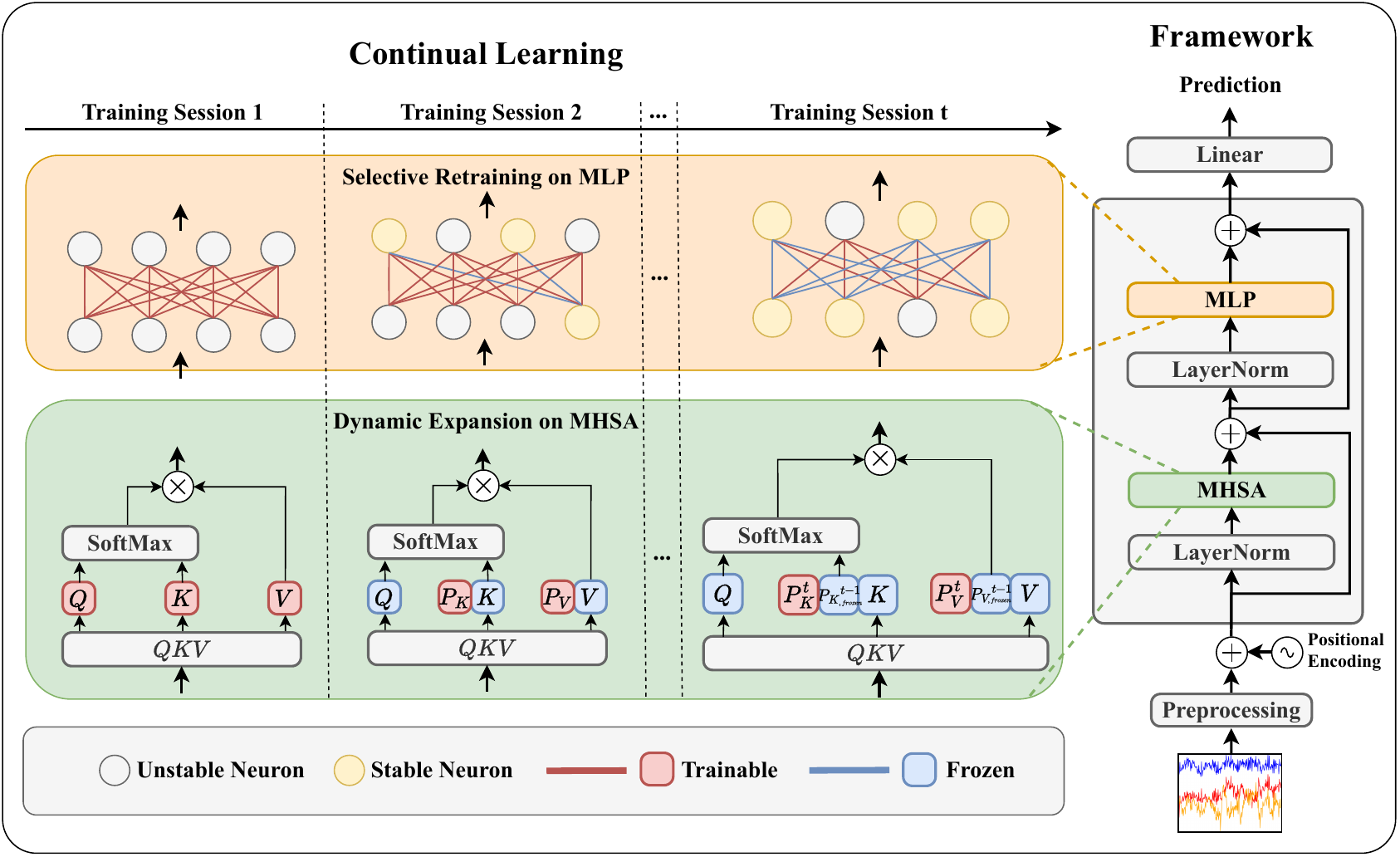} 
\caption{Architecture of FSM. The right part contains the framework. The left part details how the model dynamically expands and selectively retrains during continual learning from training session $1$ to training session $t$.
As new tasks are introduced, the model dynamically expands with new prefixes in the MHSA layer.
In the MLP, a selective retraining strategy is implemented to adjust neuron weights, preserving learned outcomes from stable neurons while updating unstable neurons to accommodate new tasks.}
\label{fig_model_part2}
\end{figure*}

\subsection{Incremental Learning at the Edge Side}
The proposed framework of FSM employs a transformer-based architecture, as shown in Fig. \ref{fig_model_part2}, which mainly consists of three components: positional encoding layer, MHSA layer, and MLP layer.
\subsubsection{Positional Encoding}
Temporal sequence modeling plays a pivotal role in distinguishing reversible activities (e.g., sitting down vs. standing up) for WiFi-based HAR.
Traditional absolute or relative position encodings assign a unique and highly distinctive code to each individual point, which may induce noise.
We use Gaussian range encoding to address this challenge, which is proposed in \cite{li2021two}.
Gaussian range encoding refers to a set of positions that share similar encoding properties based on their temporal or spatial characteristics.
This encoding allows a position to belong to multiple ranges simultaneously, and dynamically adjusts these ranges during training.
Assume that there are $G$ ranges, and denote by $z_{ij}$ a random variable that represents the occurrence of position $i$ ($i\in \{1,2,\dots,n\}$, where $n$ represents the temporal dimension of CSI) belonging to the $j$-th range, $j\in \{1,2,\dots,G\}$, with
\begin{equation}
z_{ij} \sim N(\mu_j, \sigma_j),
\end{equation}
where $\mu_j$ and $\sigma_j$ are trainable parameters defining the $j$-th range.
Denote by $p_i$ the range distribution for position $i$, which is expressed as
\begin{equation}
p_i = \langle \frac{p^1(i)}{\zeta}, \frac{p^2(i)}{\zeta}, \dots, \frac{p^G(i)}{\zeta} \rangle,
\end{equation}
where $\zeta$ represents a normalization factor, and $p^g(i)$ represents the probability of the g-th range for position $i$.
Denote by $v_j$ the value of the $j$-th range.
Given a value vector $v = \langle v_1, v_2, \dots, v_K \rangle$, the expectation for position $i$ is $p_i v^T$.
Here, $\mu_j$, $\sigma_j$, and $v_j$ are unobserved latent variables, dynamically updated through neural network back propagation.
Denote by $\beta$ ($\beta\in \mathbb{R}^{n \times G}$) the normalized weight for the $G$ Gaussian distributions, which is expressed as
\begin{equation}
\beta = \text{softmax}(B),
\end{equation}
where each element $b_{ij}$ of matrix $B$ is expressed as
\begin{equation}
b_{ij} = -\frac{(i - \mu_j)^2}{2 \sigma_j^2} - \log(\sigma_j).
\end{equation}
Denote by $X$ the original stream.
Adding the range encoding to $X$ generates a new range-biased stream, denoted by $X'$, which is expressed as
\begin{equation}
X' = X + \beta E,
\end{equation}
where $E$ ($E\in \mathbb{R}^{G \times d}$, where $d$ represents the spatial dimension of CSI) represents a set of learnable range encodings.

\begin{figure*}[htbp]
\centering
\includegraphics[width=0.98\textwidth]{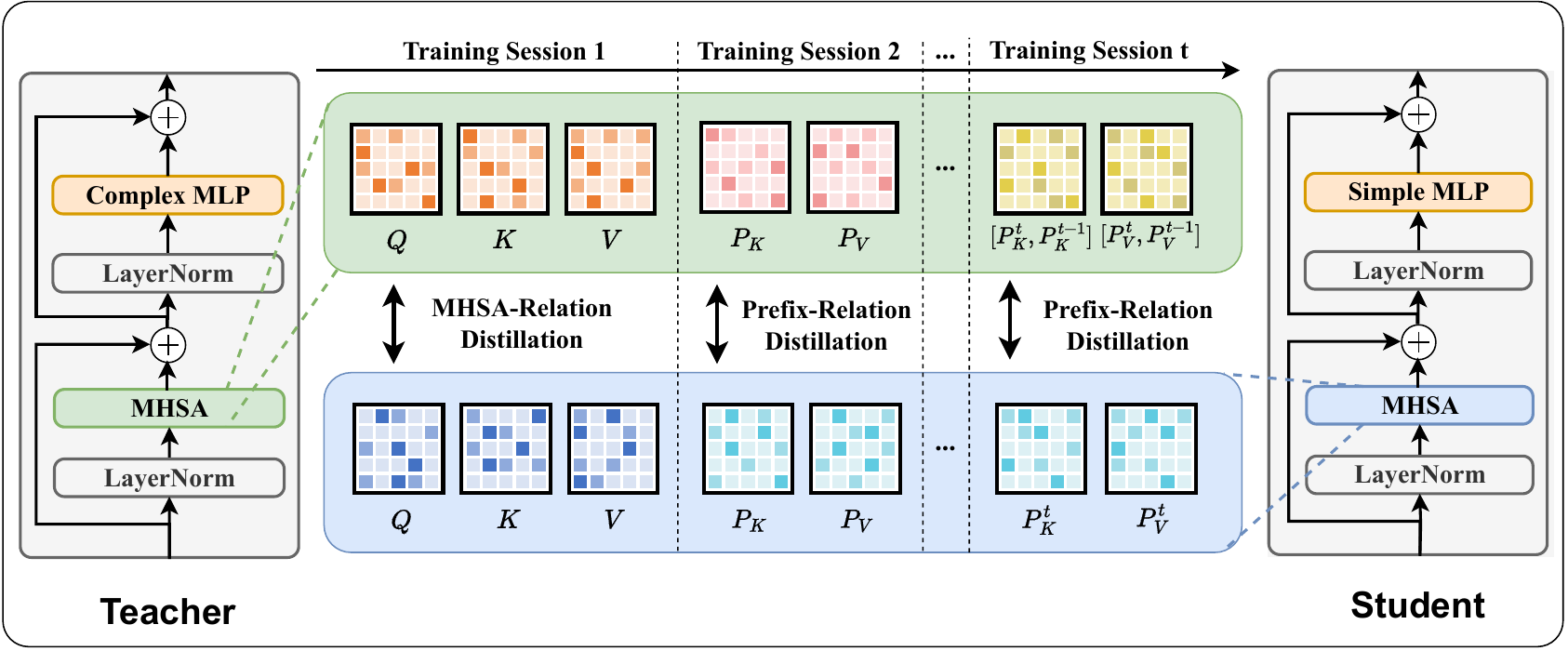} 
\caption{Overview of dual-phase distillation. The left side illustrates the teacher model, which includes complex MLP layers and MHSA layer with specific prefixes. The right side depicts the student model, which employs simplified MLP layer and MHSA layer without retaining historical prefixes. The middle section provides a detailed overview of the distillation process from the initial training stage to the t-th training stage. In the initial lightweight phase, knowledge transfer is achieved through MHSA relation distillation. In the incremental lightweight phase, prefix relation distillation compresses the teacher model’s historical prefix knowledge into the student model.}
\label{fig_model_part3}
\end{figure*}

\subsubsection{Task-specific Dynamic Expansion on MHSA}
The input sequence of MHSA $X'$, $X' \in \mathbb{R}^{n \times d}$, has been processed using Gaussian range encoding, matching the original CSI dimensions of the input.
For each attention head $h_i$, three linear transformation matrices, denoted by $W_i^Q$, $W_i^K$ and $W_i^V$, are used to transform the input $X'$ into queries (denoted by $Q$), keys (denoted by $K$) and values (denoted by $V$), respectively.
The relationship between them is expressed as
\begin{equation}
\left\{
\begin{aligned}
Q_i &= X' W_i^Q, \\
K_i &= X' W_i^K, \\
V_i &= X' W_i^V.
\end{aligned}
\right.
\end{equation}
Each attention head subsequently employs the scaled dot-product attention mechanism to calculate the corresponding attention weights.
Specifically, by computing the dot product between the queries and keys, applying softmax to obtain the weights, and then using these weights to perform a weighted sum of the values, the final attention output $\text{Attention}()$ is produced.
The formula is:
\begin{equation}\label{eq_attention}
\text{Attention}(Q_i, K_i, V_i) = \text{softmax}\left( \frac{Q_i K_i^T}{\sqrt{\frac{d}{h}}} \right) V_i,
\end{equation}
where $\frac{1}{\sqrt{\frac{d}{h}}}$ is the scaling factor to avoid large inner product values that could cause gradient vanishing issues.
Finally, the results from all heads are concatenated, and then transformed using the linear transformation matrix $W_O$, $W_O \in \mathbb{R}^{d \times d}$, to derive the final output, denoted by $\text{MultiHead}(Q, K, V)$, which is expressed as
\begin{equation}
\text{MultiHead}(Q, K, V) = [\text{head}_1, \dots, \text{head}_h] W_O,
\end{equation}
where $\text{head}_i = \text{Attention}(Q_i, K_i, V_i)$.

The objective of acquiring new knowledge while retaining old knowledge without storing exemplars can be achieved by introducing task-specific prefixes to the MHSA layers.
This approach facilitates knowledge transfer between tasks while minimally altering the model's original parameters.
By keeping the model's parameters fixed and updating only the prefixes, the system effectively prevents catastrophic forgetting while preserving adaptability. 
Specifically, each MHSA layer comprises $h$ attention heads, and the addition of prefixes to these layers enables class-incremental learning for new tasks.
Concretely, we have
\begin{equation}
\left\{
\begin{aligned}
{W_i^K}' = [P^t_K, {P^{t-1}_{K,frozen}}, W_{frozen}^K],\\
{W_i^V}' = [P^t_V, {P^{t-1}_{V,frozen}}, W_{frozen}^V].
\end{aligned}
\right.
\end{equation}

The output of head $i$ in the self-attention layer is expressed as
\begin{equation}
\text{head}_i = \text{Attention}(Q_i, {K_i}', {V_i}'), 
\end{equation}
where ${Q_i} = X'{W_{frozen}^Q}$, ${K_i}' = X'{W_i^K}'$, and ${V_i}' = X'{W_i^V}'$.
To effectively integrate prior knowledge with new information, the model sequentially concatenates the new trainable prefixes $P^t_K$ and $P^t_V$ with the previously frozen prefixes $P^{t-1}_{K,frozen}$ and $P^{t-1}_{V,frozen}$.
These concatenated prefixes are then merged with the consistently frozen weights $W_{frozen}^K$ and $W_{frozen}^V$.
This concatenation ensures a seamless transition and retention of learned features across tasks.

While prefixes are effective for class-incremental learning tasks, their random initialization can lead to unstable performance due to the variability in the initial weights.
To address this issue, we draw inspiration from the parallel attention design \cite{yu2022towards}, which employs a parallel adapter to stabilize the prefixes. 
Specifically, given an input sequence $X'$, the prefix generation is expressed as
\begin{equation}
P = Adapter(X') = Tanh(X' W_{down})W_{up}, 
\end{equation}
where $P$ represents all prefixes, $Tanh$ represents the activation function, and $W_{down}$ and $W_{up}$ denote the parameters of the parallel adapter's scaling layers. Specifically $W_{down}$ denotes a linear transformation layer that reduces the dimensionality of $X'$, and $W_{up}$ denotes another linear transformation that expands the transformed output.

\subsubsection{Stability-Aware Selective Retraining on MLP}
While the MHSA layer extracts the temporal features of CSI through linear transformations, the subsequent MLP layer enhances the representational capacity through non-linear mappings.
 To mitigate catastrophic forgetting in incremental learning, we propose a stability-aware selective retraining approach based on neuron activation for MLP layers.
This method involves three main steps, i.e., calculating each neuron's average activation value, identifying stable neurons, and generating freeze masks for parameter updates. 
This procedure is executed independently across all MLP layers, ensuring localized adaptation while preserving critical historical knowledge.

Given a training set, denote by $\bar{a}_p^{(l)}$ the average activation value for each neuron in the $l$-th layer, which is expressed as
\begin{equation}
\bar{a}_p^{(l)} = \frac{1}{B} \sum_{q=1}^{B} a_p^{(q,l)}, 
\end{equation}
where $B$, $l$, and $a_p^{(q,l)}$ denote the size of the training set, the layer index, and the activation value of the $p$-th neuron in the $l$-th layer for the $q$-th sample, respectively.
Then, by comparing the current activation values with those from the previous task, denote by $S^{(l)}$ the set of stable neurons in each layer, which is expressed as
\begin{equation}
S^{(l)} = \{ p \mid \| \bar{a}_p^{(l,t)} - \bar{a}_p^{(l,t-1)} \|_2 \leq \epsilon \}, 
\end{equation}
where $S^{(l)}$ accumulates stable neurons across the entire training period, $\epsilon$ denotes a predefined threshold, and $\bar{a}_p^{(l,t)}$ and $\bar{a}_p^{(l,t-1)}$ represent the average activation values for the current and previous tasks in the $l$-th layer, respectively.
Finally, we generate the freeze mask set $M^{(l)} = \{ M_W^{(l)}, M_b^{(l)} \}$ based on the set of stable neurons $S^{(l)}$ for each MLP layer, where $M_W^{(l)}$ and $M_b^{(l)}$ denote mask matrices corresponding to the weight matrix $W^{(l)}$ and bias vector $b^{(l)}$ in the $l$-th layer, respectively.
Matrices $M_W^{(l)}$ and $M_b^{(l)}$ are initialized with values set to one. 
Subsequently, for stable neurons in the set $S^{(l)}$, the corresponding values in $M_W^{(l)}$ and $M_b^{(l)}$ are set to zero.

During backpropagation, these masks are applied across all layers by identifying positions where $M_W^{(l)}$ and $M_b^{(l)}$ have a value of zero.
At these positions, the corresponding gradients of $W^{(l)}$ and $b^{(l)}$ are set to zero, ensuring that these parameters are not updated. Parameters that are not frozen continue to be updated normally.
Specifically, for the weight matrix $W$, the gradient update rule is expressed as
\begin{equation}
\nabla W_{ij} =
\begin{cases}
0, & \text{if } (i, j) \in M^{(l)}, \\
\nabla W_{ij}, & \text{otherwise}.
\end{cases}
\end{equation}
For the bias vector $b$, the gradient update rule is expressed as
\begin{equation}
\nabla b_i =
\begin{cases}
0, & \text{if } i \in M^{(l)}, \\
\nabla b_i, & \text{otherwise}.
\end{cases}
\end{equation}
In this manner, the approach reduces forgetting by preserving the weights of stable neurons while preventing excessive computational load during new task learning.

\subsection{Lightweight at Edge Side}
Knowledge distillation is one of the model lightweighting techniques, which can transfer the knowledge from a large teacher model $Tea$ to a smaller student model $Stu$.
The student model is trained to emulate the behavior of the teacher model, where the behavior functions of the teacher and student models are represented as $f_{Tea}$ and $f_{Stu}$, respectively.
These functions are responsible for transforming the network input into meaningful representations, typically defined by the output of any layer in the network.
In transformer-based distillation, the output of the MHSA layer, the MLP layer, or other intermediate representations (such as the attention matrix $A$) can be used as the behavior function.

The fundamental objective of knowledge distillation is to minimize the representational discrepancy between the teacher and student models in feature space. 
This is formalized through the following optimization objective
\begin{equation}
\mathcal{L}_{\text{KD}} = \sum_{x \in X} L(f_{Stu}(x), f_{Tea}(x)),
\end{equation}
where $X$ represents the training dataset, $L$ denotes the loss function, $f_{Stu}(x)$ and $f_{Tea}(x)$ are the feature representations extracted by the student and teacher models for input $x$, respectively.
A prevalent choice for the loss function is mean squared error (MSE), which quantifies the Euclidean distance between feature vectors, and is expressed as
\begin{equation}
\text{MSE}(f_{Stu}(x), f_{Tea}(x)) = \| f_{Stu}(x) - f_{Tea}(x) \|_2^2,
\end{equation}
where $\| \cdot \|_2$ denotes the Euclidean norm.
By minimizing this loss function, the student model will attempt to make its feature outputs as close as possible to those of the teacher model, thereby better mimicking the teacher's behavior.

During training session evolution, the teacher model accumulates multi-scale spatiotemporal features through task-specific prefixes.
Traditional single-phase distillation struggles to transfer historical task knowledge and new task features.
The attention module captures cross-timestep global dependencies, while prefixes encode distributions across tasks.
We therefore propose a dual-phase distillation mechanism, as shown in Fig. \ref{fig_model_part3}.
The initial phase consolidates fundamental spatiotemporal perception via MHSA relation distillation, followed by prefix relation distillation for task knowledge.
This strategy preserves the core transformer attention mechanism's representational generalization while at the same time compressing dynamically expanded parameters to a deployable scale.
Such dual optimization achieves lightweight deployment without compromising historical task konwledge.
The detailed implementation is as follows.

\subsubsection{MHSA Relation Distillation}
In the initial lightweight stage, the student model learns the behavior of the teacher model through attention relation distillation and value relation distillation. The goal is to transfer the knowledge of the teacher model to the student model via knowledge distillation. The first key component of the distillation process is the attention relation distillation loss $\mathcal{L}_{AT}$. In this stage, the student model aims to mimic the attention matrix of the teacher model. The attention matrix of each attention head in the teacher model is compared with the corresponding attention matrix in the student model. By using the MSE to characterize the difference between them, the goal is to ensure that the student model's attention distribution matches the teacher model as closely as possible. Denote by $\mathcal{L}_{AT} $ the attention relation distillation loss function, which is expressed as 
\begin{equation}
\mathcal{L}_{AT} = \frac{1}{h} \sum_{i=1}^{h} \text{MSE}(A_i^{Tea}, A_i^{Stu}),
\end{equation}
where $A_i^{Tea}$ and $A_i^{Stu}$ denote the attention matrices of the teacher and student models for the $i$-th attention head, respectively.

To further align intermediate representations, we introduce a value relation distillation loss, denoted by $\mathcal{L}_{VR}$, which uses the relationship between the values in the attention module to guide the student’s training. 
The value relation distillation loss function is defined as
\begin{equation}
\mathcal{L}_{VR} = \frac{1}{h} \sum_{i=1}^{h} \text{MSE}(V_i^{Tea}, V_i^{Stu}),
\end{equation}
where $V_i^{Tea}$ and $V_i^{Stu}$ denote the attention values for the teacher and student models for the $i$-th attention head, respectively.
By introducing the relationship between the values, the student model is guided to deeply emulate the self-attention behavior of the teacher model.

\subsubsection{Prefix Relation Distillation}
In the incremental lightweight stage, to reduce the parameter size and computational complexity of the student model, the MHSA layer of the student model does not store the prefix information from the previous tasks. Instead, the student model learns the knowledge obtained from the teacher model through a prefix relation distillation loss function. Denote by $\mathcal{L}_{P}$ the specific loss function expression, which is expressed as
\begin{align}
\mathcal{L}_{P} =  \text{MSE}([P^{t,{Tea}}_K, {P^{t-1,{Tea}}_{K,frozen}}], P^{t,Stu}_K)\notag
\\
+  \text{MSE}([P^{t,{Tea}}_V, {P^{t-1,{Tea}}_{V,frozen}}], P^{t,Stu}_V), 
\end{align}
where $P^{t,{Tea}}_K$ and $P^{t,{Tea}}_V$ represent the key and value prefixes from the teacher model for task $t$, and ${P^{t-1,{Tea}}_{K,frozen}}$ and ${P^{t-1,{Tea}}_{V,frozen}}$ represent the frozen key and value prefixes from the teacher model for task $T_{t-1}$, $P^{t,{Stu}}_K$ and $P^{t,{Stu}}_V$ represent the key and value prefixes in the student model for task $t$, respectively.

\subsection{Empty Frame Data Processing and Inference at End Side}
\subsubsection{Empty Frame Data Processing}
In the data collection phase, a multi-antenna WiFi receiver (end device) collects CSI in real-time.
However, due to environmental interference or signal instability, some subcarriers may exhibit missing measurements at specific time points, resulting in empty frames within the CSI sequence.
To address these gaps, we employ linear interpolation to fill the missing values. 
Specifically, for each subcarrier in matrix $\mathbf{D}$, if a value is missing at time point $t$, we estimate the missing value using the linear interpolation formula
\begin{align}
\mathbf{D}(t, i) = \mathbf{D}(t_1, i) + \frac{(\mathbf{D}(t_2, i) - \mathbf{D}(t_1, i))(t - t_1)}{(t_2 - t_1)},
\end{align}
where $t_1$ and $t_2$ are the nearest valid time points before and after $t$, and $\mathbf{D}(t_1, i)$ and $\mathbf{D}(t_2, i)$ are the corresponding CSI measurements, respectively.

\subsubsection{Inference}
During the inference phase on the end device, the deployed lightweight model identifies human activities based on sensing environments.
As the system continuously interacts with edge devices, it is capable of recognizing an expanding range of activities.

\section{Experiment Evaluation}

\subsection{Datasets and Settings}
We implement WECAR on heterogeneous hardware.
Specifically, we employ the Nvidia Jetson Nano, a compact but powerful AI development kit with a quad-core Cortex-A57 processor, and a 128-core GPU and 2GB memory \cite{kurniawan2021introduction}, as the edge device, and the ESP32C3 as the end device, respectively.
We use the existing dataset to simulate the data collected by the ESP32.
To validate the effectiveness of WECAR, we conduct experiments on three widely recognized public datasets, namely WiAR \cite{guo2019wiar}, MMFi \cite{yang2024mm}, and XRF \cite{wang2024xrf55} datasets, which offer a broader range of categories. The statistics of these datasets are summarized in Table \ref{tab_dataset}.

\begin{table}[h]
\small
\centering
\caption{Statistics of the datasets. The size of each dataset is $n \times d$.}
\begin{tabular}{c c c r r}
\toprule
Dataset & Class & Size & Train & Test \\
\midrule
WiAR & 16 & 270 × 90 & 384 & 96\\
MMFi & 27 & 10 × 342 & 2160 & 540\\
XRF  & 48 & 50 × 270 & 672 & 288\\
\bottomrule
\end{tabular}
\label{tab_dataset}
\end{table}

\begin{table*}[htbp]
\small
\centering
\setlength\tabcolsep{4pt}
\caption{Comparison of average accuracy $\bar{A}$ (\%) between WECAR and five other methods on WiAR, MMFi, and XRF datasets, for short task ($N=5$ or $N=6$) and long task ($N=8$ or $N=9$). The term "Params" refers to the initial number of  model parameters, measured in millions. None of the methods utilize real historical data for replay. R-DFCIL employs synthetic data to simulate the replay data.}
\begin{tabular}{c c c c c c c c c c c c}
\toprule
\multirow{2}{*}{Method} & \multicolumn{3}{c}{WiAR} & \multicolumn{3}{c}{MMFi} & \multicolumn{3}{c}{XRF} \\
\cmidrule(r){2-4} \cmidrule(r){5-7} \cmidrule(r){8-10} 
& Params & $N=5$ & $N=8$ & Params & $N=6$ & $N=9$ & Params & $N=5$ & $N=8$ \\
\midrule
    LWF        & 18.52M &41.25 $\pm$ 2.12& 39.76 $\pm$ 2.05 & 18.52M & 33.85 $\pm$ 1.89 & 30.69 $\pm$ 1.91 & 18.52M & 32.02 $\pm$ 1.78 & 29.76 $\pm$ 2.02\\
    PASS       & 11.32M &60.74 $\pm$ 1.92& 40.72 $\pm$ 2.12 & 11.32M & 45.54 $\pm$ 2.05 & 39.69 $\pm$ 2.04 & 11.32M & 47.57 $\pm$ 1.92 & 35.74 $\pm$ 1.92\\
    R-DFCIL    & 12.81M &59.62 $\pm$ 2.25& 58.80 $\pm$ 2.32 & 12.81M & 48.90 $\pm$ 2.10 & 47.88 $\pm$ 2.15 & 12.81M & 49.68 $\pm$ 1.84 & 44.89 $\pm$ 2.01\\
    PRD        & 11.75M &63.14 $\pm$ 1.51& 59.04 $\pm$ 2.02 & 11.75M & 53.69 $\pm$ 2.03 & 52.45 $\pm$ 2.18 & 11.75M & 54.02 $\pm$ 2.12 & 51.68 $\pm$ 2.05\\
    ConTraCon  & 3.60M  &-               & 50.76 $\pm$ 2.08 & 2.50M  & -                & 44.12 $\pm$ 2.06& 2.10M  & -                 & 40.54 $\pm$ 2.07\\
\textbf{WECAR}   & \textbf{1.33M} & \textbf{87.23 $\pm$ 2.93} & \textbf{86.97 $\pm$ 1.42} & \textbf{0.79M} & \textbf{74.46 $\pm$ 1.49} & \textbf{67.11 $\pm$ 0.50} & \textbf{0.58M} & \textbf{65.71 $\pm$ 1.37} & \textbf{64.61 $\pm$ 1.19}\\
\bottomrule
\end{tabular}
\label{tab_accuracy}
\end{table*}

1) \textbf{WiAR}: WiAR consists of 480 CSI samples, evenly distributed across 16 distinct classes.
We divided the dataset into training and testing subsets at a 4:1 ratio.
After data processing, the sample size is 270 x 90.
Additionally, we organized WiAR into two task types: short and long, with the latter having more number of tasks. The short task set includes 5 tasks: the first task covers 8 classes, while the following 4 tasks cover 2 classes each. In contrast, the long task set comprises 8 tasks, with each task consistently including 2 classes.

2) \textbf{MMFi}: MMFi comprises 2700 CSI samples, evenly distributed across 27 classes.
After data processing, the sample size is 10 x 342.
In MMFi, the short task category includes a total of 6 tasks. The first task covers 12 classes, while each of the next five tasks covers 3 classes. In contrast, the long task category consists of 9 tasks, with each task handling 3 classes.

3) \textbf{XRF}: XRF initially includes 55 classes, with 7 dedicated to dual-person actions. We exclude these dual-person classes due to our focus on single-person activities, leaving 48 classes and 960 CSI samples.
Each class contains 20 samples, with 14 samples per class allocated for training and the remaining 6 used for testing.
After data processing, the sample size is 50 x 270.
In XRF, the short task category consists of 5 tasks: the first task covers 24 classes, while each of the subsequent 4 tasks contains 6 classes. In contrast, the long task category is organized into 8 tasks, each responsible for analyzing 6 classes.

\subsection{Baselines}

We compare FSM with five existing EFCIL methods:

1) \textbf{LWF} \cite{li2017learning}: LWF applies knowledge distillation to prevent forgetting old tasks. It uses the original model's predictions as soft targets during new task training.

2) \textbf{PASS} \cite{zhu2021prototype}: PASS reduces catastrophic forgetting in incremental learning by combining prototype augmentation and self-supervised learning, effectively retaining past knowledge while adapting to new tasks.

3) \textbf{R-DFCIL} \cite{gao2022r}: R-DFCIL synthesizes data for previous classes using model inversion and applies relation-guided representation learning to minimize the domain gap between synthetic and real data.

4) \textbf{PRD} \cite{asadi2023prototype}: PRD introduces a new distillation loss to maintain the relevance of class prototypes during new task learning, ensuring that the current model maintains relevance to the prototypes of previous tasks while adapting to new tasks.

5) \textbf{ConTraCon} \cite{roy2023exemplar}: ConTraCon modifies the MHSA layer weights via convolutional operations to adapt the transformer architecture for new tasks.

\subsection{Evaluation Metrics}
We use two metrics, i.e., the average accuracy and average forgetting to measure the performance of WECAR.
The accuracy after each task, denoted by $A_t$, represents the accuracy over all classes learned up to and including the $t$-th task. Subsequently, the average accuracy across all tasks, represented by $\bar{A}$, is expressed as
\begin{equation}
\bar{A} = \frac{1}{N} \sum_{t=1}^{N} A_t,
\notag
\end{equation}
where $N$ represents the number of tasks.
The average forgetting measure \cite{chaudhry2018riemannian} is used to estimate the forgetting of previous tasks.
For each task $t$, the forgetting measure of predicting previous task $k$ is denoted by $f_k^t$, which is expressed as
\begin{equation}
f_k^t = \max_{z \in \{1, \ldots, k-1\}} (\alpha_{z,t} - \alpha_{k,t}),
\notag
\end{equation}
where $\alpha_{m,j}$ represents the accuracy of task $j$ after training task $m$.  The average forgetting measure represents the forgetting measure of the last task, denoted by $\bar{F}$, which is expressed as
\begin{equation}
{\bar{F}} = \frac{1}{N-1} \sum_{k=1}^{N-1} f_k^N.
\notag
\end{equation}

\subsection{Implementation Details }

Our method is implemented by PyTorch \cite{paszke2019pytorch} and trained on Nvidia Jetson Nano GPU with 2GB memory.
The optimizer chosen is Adam \cite{kingma2014adam}, with an initial learning rate of 0.001 and a batch size of 4.
The model's training cycle is set to 50 epochs.
After completing the lightweighting process, the model is converted into the ONNX format.
Subsequently, the ONNX model is processed to a binary file, which is transferred to the ESP32C3 and flashed onto the device to execute inference.

During the model training process, we set the number of Gaussian distributions in the positional encoding to 10.
The values of $\mu$s are uniformly distributed across the temporal dimension for various datasets as follows.
For WiAR, they range from 13.5 to 256.5 with a step size of 27.
For MMFi, they range from 0.5 to 9.5 with a step size of 1.
For XRF, they range from 2.5 to 47.5 with a step size of 5.
The standard deviation of the Gaussian distributions on all the datasets is uniformly set to 8.
The number of stacks in the module is set to 1.
The input dimensions for the three datasets are set to 90, 342, and 270, respectively, while 
maintaining a consistent number of heads at 9 for each, and employing a dropout rate of 0.1.

\begin{table}[h]

\small
\centering
\setlength{\tabcolsep}{3pt} 
\caption{Comparison of the average forgetting measure $\bar{F}$ (\%) between WECAR and five other methods on the WiAR, MMFi, and XRF datasets, with short and long tasks (lower values indicate better performance).}
\begin{tabular}{c c c c c c c}
\toprule
\multirow{2}{*}{Method} & \multicolumn{2}{c}{WiAR} & \multicolumn{2}{c}{MMFi} & \multicolumn{2}{c}{XRF} \\
\cmidrule(r){2-3} \cmidrule(r){4-5} \cmidrule(r){6-7}
& $N=5$ & $N=8$ & $N=6$ & $N=9$ & $N=5$ & $N=8$ \\
\midrule
                LWF           & 30.32 & 30.21 & 31.05 & 32.57 & 33.53 & 30.01\\
                PASS          & 24.14 & 29.34 & 22.02 & 25.17 & 21.09 & 29.03\\
                R-DFCIL       & 23.24 & 23.96 & 20.98 & 24.61 & 21.35 & 26.83\\
                PRD           & 20.90 & 22.02 & 19.81 & 19.24 & 24.97 & 20.69\\
                ConTraCon     & -     & 27.56 & -     & 28.98 & -     & 25.86\\
\textbf{WECAR}      & \textbf{15.36} & \textbf{13.19} & \textbf{17.94} & \textbf{19.09} & \textbf{20.29} & \textbf{18.69}\\
\bottomrule
\end{tabular}
\label{tab_forgetting}
\end{table}

\pgfplotscreateplotcyclelist{colorlist1}{%
    {blue, mark=*},
    {brown, mark=*},
    {green, mark=*},
    {orange, mark=*},
    {red, mark=triangle*},
}
\pgfplotscreateplotcyclelist{colorlist2}{%
    {blue, mark=*},
    {brown, mark=*},
    {green, mark=*},
    {orange, mark=*},
    {purple, mark=*},
    {red, mark=triangle*},
}

\begin{figure*}[htbp]
\centering
\includegraphics[width=0.86\textwidth]{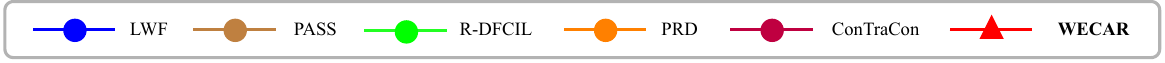} 
\label{fig_legend}
\vspace{-0.35cm} 
\end{figure*}

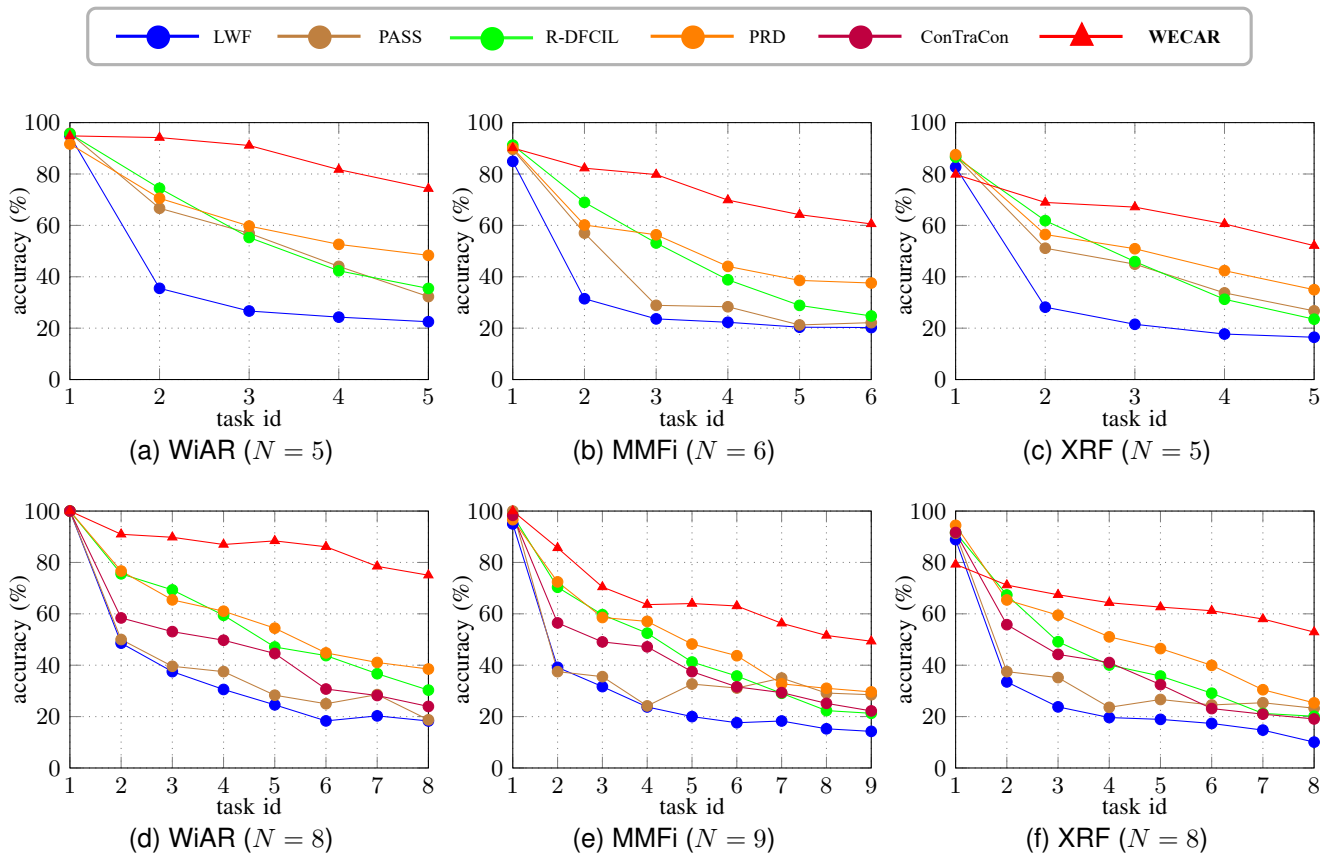
\begin{figure*}[!htb]
    \centering
    \subfloat[WiAR ($N=5$)]{%
        \begin{tikzpicture}
        \begin{axis}[
            width=0.35\linewidth,
            height=5cm,
            xlabel={task id},
            ylabel={accuracy (\%)},
            cycle list name=colorlist1,
            label style={font=\small, inner sep=0pt},
            tick label style={font=\small},
            grid=major,
            grid style={dotted, gray},
            legend pos=south west,
            legend style={fill=none, font=\fontsize{4}{6}\selectfont,legend columns=3, draw=none},
            xlabel style={font=\small, at={(axis description cs:0.5,0.04)}},
            ylabel style={font=\small, at={(axis description cs:0.12,0.5)}},
            xmin=1, ymin=0, 
            xmax=5, ymax=100,
            xtick={1, 2, 3, 4, 5}, 
            ytick={0, 20, 40, 60, 80, 100}, 
        ]
        \addplot coordinates {(1,94.92) (2,35.50) (3,26.67) (4,24.29) (5,22.50)};
        \addplot coordinates {(1,95.83) (2,66.67) (3,56.94) (4,44.04) (5,32.29)};
        \addplot coordinates {(1,95.65) (2,74.43) (3,55.29) (4,42.34) (5,35.46)};
        \addplot coordinates {(1,91.67) (2,70.56) (3,59.74) (4,52.61) (5,48.34)};
        \addplot coordinates {(1,94.83) (2,94.17) (3,91.10) (4,81.72) (5,74.30)};
        \end{axis}
        \end{tikzpicture}
    }
    \subfloat[MMFi ($N=6$)]{%
        \begin{tikzpicture}
        \begin{axis}[
            width=0.35\linewidth,
            height=5cm,
            xlabel={task id},
            ylabel={accuracy (\%)},
            cycle list name=colorlist1,
            label style={font=\small, inner sep=0pt},
            tick label style={font=\small},
            grid=major,
            grid style={dotted, gray},
            legend pos=south west,
            legend style={fill=none, font=\fontsize{4}{10}\selectfont,legend columns=3, draw=none},
            xlabel style={font=\small, at={(axis description cs:0.5,0.04)}},
            ylabel style={font=\small, at={(axis description cs:0.12,0.5)}},
            xmin=1, ymin=0, 
            xmax=6, ymax=100,
            xtick={1, 2, 3, 4, 5,6}, 
            ytick={0, 20, 40, 60, 80, 100}, 
        ]
        \addplot coordinates {(1,84.92) (2,31.45) (3,23.61) (4,22.29) (5,20.37) (6,20.24)};
        \addplot coordinates {(1,89.58) (2,57.00) (3,28.88) (4,28.33) (5,21.25) (6,22.14)};
        \addplot coordinates {(1,91.25) (2,68.98) (3,53.14) (4,38.83) (5,28.86) (6,24.74)};
        \addplot coordinates {(1,90.08) (2,60.15) (3,56.34) (4,44.04) (5,38.58) (6,37.57)};
        \addplot coordinates {(1,90.13) (2,82.26) (3,79.78) (4,69.84) (5,64.17) (6,60.57)};
        \end{axis}
        \end{tikzpicture}
    }
    \subfloat[XRF ($N=5$)]{%
        \begin{tikzpicture}
        \begin{axis}[
            width=0.35\linewidth,
            height=5cm,
            xlabel={task id},
            ylabel={accuracy (\%)},
            cycle list name=colorlist1,
            label style={font=\small, inner sep=0pt},
            tick label style={font=\small},
            grid=major,
            grid style={dotted, gray},
            legend pos=south west,
            legend style={fill=none, font=\fontsize{4}{10}\selectfont,legend columns=3, draw=none},
            xlabel style={font=\small, at={(axis description cs:0.5,0.04)}},
            ylabel style={font=\small, at={(axis description cs:0.12,0.5)}},
            xmin=1, ymin=0, 
            xmax=5, ymax=100,
            xtick={1, 2, 3, 4, 5}, 
            ytick={0, 20, 40, 60, 80, 100}, 
        ]
        \addplot coordinates {(1,82.66) (2,28.17) (3,21.50) (4,17.71) (5,16.46)};
        \addplot coordinates {(1,86.80) (2,51.11) (3,44.90) (4,33.73) (5,26.73)};
        \addplot coordinates {(1,86.59) (2,61.84) (3,45.91) (4,31.25) (5,23.49)};
        \addplot coordinates {(1,87.51) (2,56.45) (3,50.91) (4,42.38) (5,34.97)};
        \addplot coordinates {(1,79.82) (2,68.93) (3,67.11) (4,60.58) (5,52.11)};
        \end{axis}
        \end{tikzpicture}
    }
    \\
    \subfloat[WiAR ($N=8$)]{%
        \begin{tikzpicture}
        \begin{axis}[
            width=0.35\linewidth,
            height=5cm,
            xlabel={task id},
            ylabel={accuracy (\%)},
            cycle list name=colorlist2,
            label style={font=\small, inner sep=0pt},
            tick label style={font=\small},
            grid=major,
            grid style={dotted, gray},
            legend pos=south west,
            legend style={fill=none, font=\fontsize{6}{10}\selectfont,legend columns=3, draw=none,at={(-0.01, -0.03)}},
            xlabel style={font=\small, at={(axis description cs:0.5,0.04)}},
            ylabel style={font=\small, at={(axis description cs:0.12,0.5)}},
            xmin=1, ymin=0, 
            xmax=8, ymax=100,
            xtick={1, 2, 3, 4, 5,6,7,8}, 
            ytick={0, 20, 40, 60, 80, 100}, 
        ]
        \addplot coordinates {(1,100.00) (2,48.59) (3,37.50) (4,30.56) (5,24.58) (6,18.33) (7,20.28) (8,18.29)};
        \addplot coordinates {(1,100.00) (2,50.00) (3,39.58) (4,37.50) (5,28.33) (6,25.00) (7,28.57) (8,18.75)};
        \addplot coordinates {(1,100.00) (2,75.56) (3,69.31) (4,59.43) (5,47.09) (6,43.74) (7,36.70) (8,30.31)};
        \addplot coordinates {(1,100.00) (2,76.65) (3,65.43) (4,61.02) (5,54.42) (6,44.79) (7,41.07) (8,38.51)};
        \addplot coordinates {(1,100.00) (2,58.40) (3,53.07) (4,49.72) (5,44.57) (6,30.73) (7,28.21) (8,23.95)};
        \addplot coordinates {(1,100.00) (2,90.96) (3,89.79) (4,86.98) (5,88.39) (6,86.11) (7,78.48) (8,75.03)};
        \end{axis}
        \end{tikzpicture}
    }
    \subfloat[MMFi ($N=9$)]{%
        \begin{tikzpicture}
        \begin{axis}[
            width=0.35\linewidth,
            height=5cm,
            xlabel={task id},
            ylabel={accuracy (\%)},
            cycle list name=colorlist2,
            label style={font=\small, inner sep=0pt},
            tick label style={font=\small},
            grid=major,
            grid style={dotted, gray},
            legend pos=north east,
            legend style={fill=none, font=\fontsize{5}{10}\selectfont,legend columns=2, draw=gray!40,at={(0.96, 0.95)}},
            xlabel style={font=\small, at={(axis description cs:0.5,0.04)}},
            ylabel style={font=\small, at={(axis description cs:0.12,0.5)}},
            xmin=1, ymin=0, 
            xmax=9, ymax=100,
            xtick={1, 2, 3, 4, 5,6,7,8,9}, 
            ytick={0, 20, 40, 60, 80, 100}, 
        ]
        \addplot coordinates {(1,95.00) (2,39.17) (3,31.67) (4,23.75) (5,20.00) (6,17.61) (7,18.29) (8,15.25) (9,14.25)};
        \addplot coordinates {(1,100.00) (2,37.50) (3,35.55) (4,24.16) (5,32.66) (6,31.11) (7,35.00) (8,29.16) (9,28.51)};
        \addplot coordinates {(1,98.33) (2,70.34) (3,59.67) (4,52.48) (5,41.24) (6,35.79) (7,29.07) (8,22.34) (9,21.28)};
        \addplot coordinates {(1,96.66) (2,72.43) (3,58.57) (4,57.05) (5,48.24) (6,43.70) (7,32.75) (8,31.02) (9,29.61)};
        \addplot coordinates {(1,98.33) (2,56.46) (3,49.07) (4,47.15) (5,37.49) (6,31.57) (7,29.33) (8,25.16) (9,22.21)};
        \addplot coordinates {(1,100.00) (2,85.66) (3,70.44) (4,63.58) (5,64.00) (6,63.05) (7,56.33) (8,51.58) (9,49.33)};
        \end{axis}
        \end{tikzpicture}
    }
    \subfloat[XRF ($N=8$)]{%
        \begin{tikzpicture}
        \begin{axis}[
            width=0.35\linewidth,
            height=5cm,
            xlabel={task id},
            ylabel={accuracy (\%)},
            cycle list name=colorlist2,
            label style={font=\small, inner sep=0pt},
            tick label style={font=\small},
            grid=major,
            grid style={dotted, gray},
            legend pos=north east,
            legend style={fill=none, font=\fontsize{5}{10}\selectfont,legend columns=2, draw=gray!40,at={(0.96, 0.95)}},
            xlabel style={font=\small, at={(axis description cs:0.5,0.04)}},
            ylabel style={font=\small, at={(axis description cs:0.12,0.5)}},
            xmin=1, ymin=0, 
            xmax=8, ymax=100,
            xtick={1, 2, 3, 4, 5,6,7,8}, 
            ytick={0, 20, 40, 60, 80, 100}, 
        ]
        \addplot coordinates {(1,88.89) (2,33.50) (3,23.78) (4,19.63) (5,18.92) (6,17.33) (7,14.74) (8,10.07)};
        \addplot coordinates {(1,91.32) (2,37.50) (3,35.18) (4,23.61) (5,26.66) (6,24.53) (7,25.39) (8,23.26)};
        \addplot coordinates {(1,91.66) (2,67.33) (3,49.14) (4,40.07) (5,35.78) (6,29.10) (7,21.24) (8,20.13)};
        \addplot coordinates {(1,94.44) (2,65.40) (3,59.47) (4,51.04) (5,46.47) (6,39.95) (7,30.46) (8,25.34)};
        \addplot coordinates {(1,91.66) (2,55.78) (3,44.21) (4,41.04) (5,32.45) (6,23.12) (7,20.95) (8,19.09)};
        \addplot coordinates {(1,79.24) (2,71.18) (3,67.42) (4,64.32) (5,62.62) (6,61.20) (7,57.98) (8,52.89)};
        \end{axis}
        \end{tikzpicture}
    }
    \caption{Accuracy comparison between WECAR and the five other methods across each task. The x-axis represents task $t$, and the corresponding value of the y-axis is the accuracy, i.e., $A_t$.}
    \label{fig_acc} 
\end{figure*}

\subsection{Result Comparison}
\subsubsection{Performance Comparison}

Tables \ref{tab_accuracy} and \ref{tab_forgetting} summarize the results of WECAR compared to baselines in terms of the average accuracy $\bar{A}$ and the average forgetting $\bar{F}$, respectively. 
The two tables reveal that WECAR significantly outperforms the other baselines across all the datasets. 
Specifically, on the WiAR dataset with long task sequences ($N=8$ or $N=9$), the average accuracy of WECAR surpasses that of the other methods by nearly 30\%.
On the MMFi dataset with the short task sequences ($N=5$ or $N=6$), the average accuracy improvement exceeds 20\%.
Especially compared to LWF, the advantage of WECAR reaches 40\%.
In addition, WECAR achieves a forgetting rate of less than 21\% for both short and long tasks across the three datasets, and outperforms the other methods.
Notably, the baseline methods are evaluated in their original non-lightweighted forms, while WECAR adopts a lightweight design.
Even after reducing parameters through lightweighting, WECAR still performs better than the non-lightweight baselines.
This means if we would compare WECAR with the lightweight versions of the other methods, it should perform even better.
Moreover, WECAR does well in both plasticity and stability.
The reason is that 
in WECAR, we utilize MHSA and positional encoding, which are particularly well-suited to the characteristics of time-series data, such as the patterns and intensity of signal changes in CSI. 
These features pose challenges to traditional image-based network architectures, like Resnet, which primarily is optimized for spatial feature extraction and struggles with the dynamic characteristics of time-series data.
In contrast, 
other methods face limitations in adapting to the dynamic environment of CSI.
For instance, the forgetting-preventing approach of LWF performs poorly in dynamically changing environments.
R-DFCIL's synthetic data approach fails to accurately capture the true characteristics of CSI. 
PRD attempts to mitigate forgetting by maintaining relationships between class prototypes, but the high dynamics and complexity of CSI may render this prototype-based method ineffective.
ConTraCon, although using the transformer architecture to adapt to new tasks, relies heavily on its attention mechanism. 
If this mechanism fails to capture the temporal and frequency domain characteristics of CSI, its performance becomes suboptimal.
Moreover, its entropy-based task prediction, which relies on image enhancement techniques, is unsuitable for CSI, as temporal delays and frequency shifts in CSI do not correspond to visual changes.

\subsubsection{Comparison of Model Parameter Counts}
As shown in Table \ref{tab_accuracy}, WECAR demonstrates a significant reduction in model parameters compared to the other methods. 
Specifically, WECAR requires only one-third of the parameter count of ConTraCon and at least eight times fewer parameters than the other methods.
Notably, on the MMFi and XRF datasets, WECAR's parameter count is as low as one-tenth of those of the other methods, excluding ConTraCon. 
This advantage indicates that WECAR is particularly well-suited for deployment on edge devices, such as WiFi-based HAR terminal.

\pgfplotscreateplotcyclelist{colorlist4}{%
    {red, mark=*},
    {orange, mark=*},
    {green, mark=*},
}

\begin{figure*}[!htb]
    \centering
    \subfloat[WiAR ($N=5$)]{%
        \begin{tikzpicture}
        \begin{axis}[
            width=0.35\linewidth,
            height=5cm,
            xlabel={task id},
            ylabel={accuracy (\%)},
            cycle list name=colorlist4,
            label style={font=\small, inner sep=0pt},
            tick label style={font=\small},
            grid=major,
            grid style={dotted, gray},
            legend pos=south west,
            legend style={fill=none, font=\fontsize{7}{10}\selectfont,legend columns=3, draw=gray!40,at={(0.1, 0.2)}},
            xlabel style={font=\small, at={(axis description cs:0.5,0.04)}},
            ylabel style={font=\small, at={(axis description cs:0.12,0.5)}},
            xmin=1, ymin=60, 
            xmax=5, ymax=100,
            xtick={1, 2, 3, 4, 5}, 
            ytick={60, 80, 100}, 
        ]
        \addlegendentry{CFM}
        \addplot coordinates {(1,97.92) (2,97.29) (3,93.89) (4,86.67) (5,80.83)};
        \addlegendentry{WECAR}
        \addplot coordinates {(1,94.83) (2,94.17) (3,91.10) (4,81.72) (5,74.30)};
        
        \end{axis}
        \end{tikzpicture}
    }
    \subfloat[MMFi ($N=6$)]{%
        \begin{tikzpicture}
        \begin{axis}[
            width=0.35\linewidth,
            height=5cm,
            xlabel={task id},
            ylabel={accuracy (\%)},
            cycle list name=colorlist4,
            label style={font=\small, inner sep=0pt},
            tick label style={font=\small},
            grid=major,
            grid style={dotted, gray},
            legend pos=south west,
            legend style={fill=none, font=\fontsize{4}{10}\selectfont,legend columns=3, draw=none},
            xlabel style={font=\small, at={(axis description cs:0.5,0.04)}},
            ylabel style={font=\small, at={(axis description cs:0.12,0.5)}},
            xmin=1, ymin=40, 
            xmax=6, ymax=100,
            xtick={1, 2, 3, 4, 5,6}, 
            ytick={40, 60, 80, 100}, 
        ]
        \addplot coordinates {(1,92.80) (2,85.86) (3,82.74) (4,73.32) (5,66.96) (6,63.54)};
        \addplot coordinates {(1,90.13) (2,82.26) (3,79.78) (4,69.84) (5,64.17) (6,60.57)};
        
        \end{axis}
        \end{tikzpicture}
    }
    \subfloat[XRF ($N=5$)]{%
        \begin{tikzpicture}
        \begin{axis}[
            width=0.35\linewidth,
            height=5cm,
            xlabel={task id},
            ylabel={accuracy (\%)},
            cycle list name=colorlist4,
            label style={font=\small, inner sep=0pt},
            tick label style={font=\small},
            grid=major,
            grid style={dotted, gray},
            legend pos=south west,
            legend style={fill=none, font=\fontsize{4}{10}\selectfont,legend columns=3, draw=none},
            xlabel style={font=\small, at={(axis description cs:0.5,0.04)}},
            ylabel style={font=\small, at={(axis description cs:0.12,0.5)}},
            xmin=1, ymin=20, 
            xmax=5, ymax=100,
            xtick={1, 2, 3, 4, 5}, 
            ytick={20, 40, 60, 80, 100}, 
        ]
        \addplot coordinates {(1,86.70) (2,72.17) (3,68.79) (4,65.52) (5,56.34)};
        \addplot coordinates {(1,79.82) (2,68.93) (3,67.11) (4,60.58) (5,52.11)};
        
        \end{axis}
        \end{tikzpicture}
    }
    \\
    \subfloat[WiAR ($N=8$)]{%
        \begin{tikzpicture}
        \begin{axis}[
            width=0.35\linewidth,
            height=5cm,
            xlabel={task id},
            ylabel={accuracy (\%)},
            cycle list name=colorlist4,
            label style={font=\small, inner sep=0pt},
            tick label style={font=\small},
            grid=major,
            grid style={dotted, gray},
            legend pos=south west,
            legend style={fill=none, font=\fontsize{6}{10}\selectfont,legend columns=3, draw=none,at={(-0.01, -0.03)}},
            xlabel style={font=\small, at={(axis description cs:0.5,0.04)}},
            ylabel style={font=\small, at={(axis description cs:0.12,0.5)}},
            xmin=1, ymin=60, 
            xmax=8, ymax=100,
            xtick={1, 2, 3, 4, 5,6,7,8}, 
            ytick={60, 80, 100}, 
        ]
        \addplot coordinates {(1,100.00) (2,94.15) (3,92.50) (4,88.76) (5,89.46) (6,88.50) (7,85.65) (8,80.71)};
        \addplot coordinates {(1,100.00) (2,90.96) (3,89.79) (4,86.98) (5,88.39) (6,86.11) (7,78.48) (8,75.03)};
        
        \end{axis}
        \end{tikzpicture}
    }
    \subfloat[MMFi ($N=9$)]{%
        \begin{tikzpicture}
        \begin{axis}[
            width=0.35\linewidth,
            height=5cm,
            xlabel={task id},
            ylabel={accuracy (\%)},
            cycle list name=colorlist4,
            label style={font=\small, inner sep=0pt},
            tick label style={font=\small},
            grid=major,
            grid style={dotted, gray},
            legend pos=north east,
            legend style={fill=none, font=\fontsize{5}{10}\selectfont,legend columns=2, draw=gray!40,at={(0.96, 0.95)}},
            xlabel style={font=\small, at={(axis description cs:0.5,0.04)}},
            ylabel style={font=\small, at={(axis description cs:0.12,0.5)}},
            xmin=1, ymin=40, 
            xmax=9, ymax=100,
            xtick={1, 2, 3, 4, 5,6,7,8,9}, 
            ytick={40, 60, 80, 100}, 
        ]
        \addplot coordinates {(1,100.00) (2,92.16) (3,78.33) (4,65.16) (5,65.77) (6,64.00) (7,58.18) (8,55.46) (9,51.46)};
        \addplot coordinates {(1,100.00) (2,85.66) (3,70.44) (4,63.58) (5,64.00) (6,63.05) (7,56.33) (8,51.58) (9,49.33)};
        
        \end{axis}
        \end{tikzpicture}
    }
    \subfloat[XRF ($N=8$)]{%
        \begin{tikzpicture}
        \begin{axis}[
            width=0.35\linewidth,
            height=5cm,
            xlabel={task id},
            ylabel={accuracy (\%)},
            cycle list name=colorlist4,
            label style={font=\small, inner sep=0pt},
            tick label style={font=\small},
            grid=major,
            grid style={dotted, gray},
            legend pos=north east,
            legend style={fill=none, font=\fontsize{5}{10}\selectfont,legend columns=2, draw=gray!40,at={(0.96, 0.95)}},
            xlabel style={font=\small, at={(axis description cs:0.5,0.04)}},
            ylabel style={font=\small, at={(axis description cs:0.12,0.5)}},
            xmin=1, ymin=20, 
            xmax=8, ymax=100,
            xtick={1, 2, 3, 4, 5,6,7,8}, 
            ytick={20, 40, 60, 80, 100}, 
        ]
        \addplot coordinates {(1,88.33) (2,78.93) (3,73.64) (4,66.63) (5,61.27) (6,62.19) (7,57.92) (8,54.02)};
        \addplot coordinates {(1,79.24) (2,71.18) (3,67.42) (4,64.32) (5,62.62) (6,61.20) (7,57.98) (8,52.89)};
        
        \end{axis}
        \end{tikzpicture}
    }
    \caption{Accuracy comparison of WECAR and CFM across each task. The x-axis represents task $t$, and the corresponding value of the y-axis is the accuracy, i.e., $A_t$.}
    \label{fig_three_acc} 
\end{figure*}
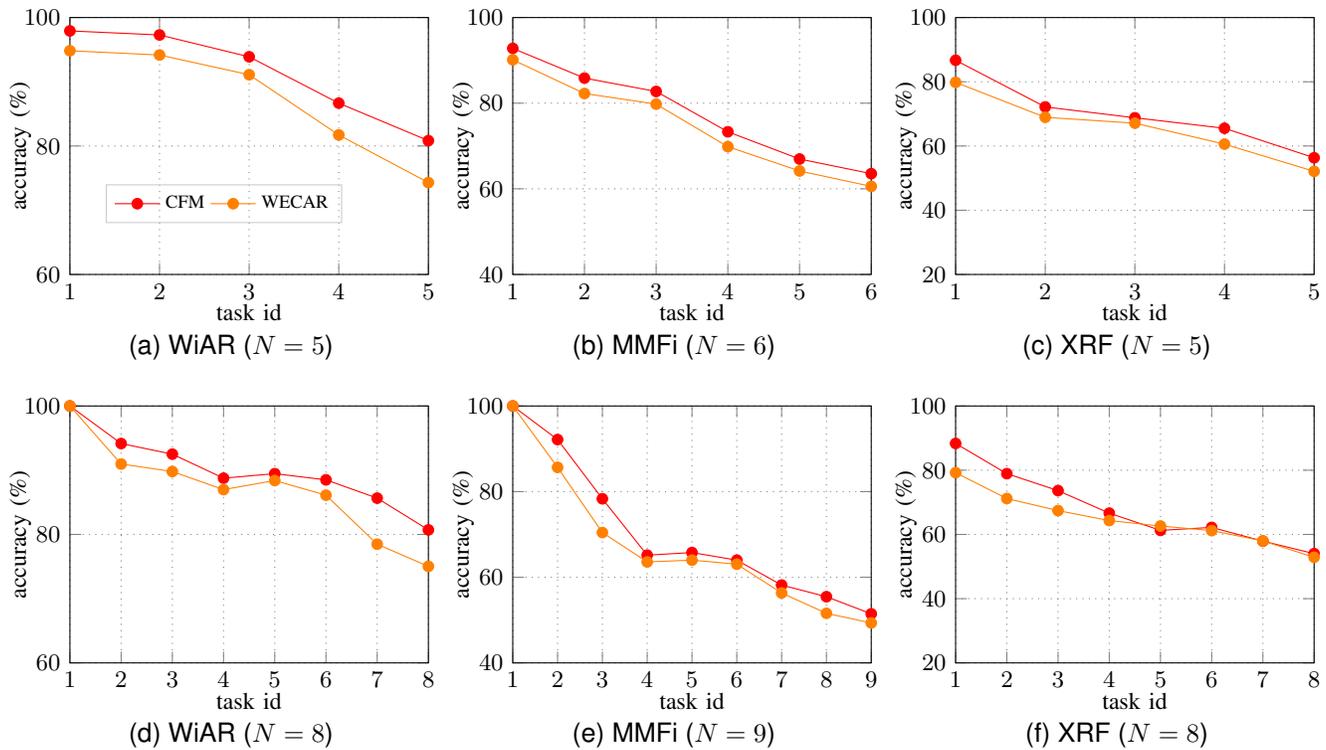

\subsubsection{Comparison of Accuracy for Each Task}
In Fig. \ref{fig_acc}, we compare the accuracy of WECAR with other methods across each task on the three datasets. 
We observe that the accuracy for the initial task is generally comparable among all methods, with the exception of WECAR, which exhibits slightly lower performance on the short and long tasks of the XRF dataset. 
However, the accuracy of WECAR significantly outperforms the other methods in subsequent tasks. 
This demonstrates that our method achieves a balance between knowledge forgetting and acquisition when dealing with CSI. 
In addition, the performance gap between WECAR and the other methods widens as the number of tasks increases.
For example, in Fig. \ref{fig_acc}(e), 
at the fifth task, the gain of WECAR compared to PRD is approximately 10\%, while by the ninth task, the improvement grows to 20\%.
This trend demonstrates the effectiveness of WECAR in handling extended task sequences, highlighting its robustness in continually sensing.

\begin{table}[htbp]
\small
\centering
\caption{Comparison of model metrics with and without dual-phase distillation. 
"Size" indicates the model file storage, quantified in megabytes (MB). "GPU" denotes the peak memory consumption during training, measured in MB.
Notably, the GPU metric for WECAR corresponds to its memory usage without employing CFM as a teacher model in dual-phase distillation.}
\begin{tabular}{c c c c}
\toprule
Dataset & & CFM & WECAR \\
\midrule
\multirow{5}{*}{WiAR}        & Params & 3.35M  & 1.33M      \\
                             & Size   & 12.47MB  & 4.95MB      \\
                             & GPU    & 342.55MB  & 257.67MB      \\
                             & $N=5$  & 91.32 $\pm$ 2.22  & 87.23 $\pm$ 2.93\\
                             & $N=8$  & 89.97 $\pm$ 1.31  & 86.97 $\pm$ 1.42\\
\midrule
\multirow{5}{*}{MMFi}        & Params & 1.92M  & 0.79M      \\
                             & Size   & 7.15MB  & 3.03MB      \\
                             & GPU    & 20.99MB  & 15.23MB      \\
                             & $N=6$  & 77.54 $\pm$ 2.02  & 74.46 $\pm$ 1.49\\
                             & $N=9$  & 70.06 $\pm$ 1.41  & 67.11 $\pm$ 0.50\\
\midrule
\multirow{5}{*}{XRF}         & Params & 1.50M  & 0.58M      \\
                             & Size   & 5.59MB  & 2.16MB      \\
                             & GPU    & 94.09MB  & 67.18MB      \\
                             & $N=5$  & 69.91 $\pm$ 1.42  & 65.71 $\pm$ 1.37\\
                             & $N=8$  & 67.87 $\pm$ 1.65  & 64.61 $\pm$ 1.19\\
\bottomrule
\end{tabular}
\label{tab_distill}
\end{table}

\subsection{Ablation Test }



\subsubsection{Effect of Dual-Phase Distillation}
Fig. \ref{fig_three_acc} illustrates the task-wise accuracy comparison with and without dual-phase distillation, while Table \ref{tab_distill} presents the model metrics and average accuracy comparison.
Here, CFM denotes the results without dual-phase  distillation, which maintains complete parameter integrity during continual learning.
It can be observed that WECAR significantly reduces parameters, model size, and GPU memory usage while maintaining high classification accuracy.
Across the WiAR, MMFi, and XRF datasets, WECAR achieves a 60\%-70\% reduction in parameter count and around a 25\% decrease in GPU memory consumption compared to CFM, with a classification accuracy dropping of only 2\%-4\% for short and long tasks.
These findings highlight the effectiveness of the proposed dual-phase distillation in achieving substantial model simplification and resource efficiency, making WECAR well-suited for deployment on resource-constrained edge devices.


\begin{table}[h]
\small
\centering
\caption{Ablation study of two strategies for long tasks across the three datasets. Strategy 1 represents dynamic expansion on MHSA. Strategy 2 represents selective retraining on MLP.}
\begin{tabular}{c c c c c}
\toprule
Dataset & Strategy 1 & Strategy 2 & $A_N$ & $\bar{A}$ \\
\midrule
\multirow{4}{*}{WiAR} & $\times$ & $\times$ & 36.45 & 52.08 \\
                      & $\checkmark$ & $\times$ & 51.03 & 67.70 \\
                      & $\times$ & $\checkmark$ & 44.79 & 59.37 \\
                      & $\checkmark$ & $\checkmark$ & \textbf{78.58} & \textbf{89.85} \\
\midrule
\multirow{4}{*}{MMFi} & $\times$ & $\times$ & 25.91 & 46.27 \\
                      & $\checkmark$ & $\times$ & 38.87 & 60.15 \\
                      & $\times$ & $\checkmark$ & 34.24 & 54.78 \\
                      &  $\checkmark$ & $\checkmark$ & \textbf{53.52} & \textbf{71.97} \\
\midrule
\multirow{4}{*}{XRF} & $\times$ & $\times$ & 22.56 & 40.96 \\
                      & $\checkmark$ & $\times$ & 41.66 & 53.12 \\
                      & $\times$ & $\checkmark$ & 30.90 & 48.60 \\
                      &  $\checkmark$ & $\checkmark$ & \textbf{48.71} & \textbf{65.79} \\
\bottomrule
\end{tabular}
\label{ablation}
\end{table}

\subsubsection{Effect of Dynamic Expansion and Selective Retraining}

Table \ref{ablation} demonstrates that the two strategies, i.e., dynamic expansion on MHSA and selective retraining on MLP, significantly enhance the performance of WECAR in long-term EFCIL ablation experiments.
More specifically, dynamic expansion significantly enhances the accuracy of the last task and average task accuracy across all the datasets.
It achieves this by adding trainable prefixes to the multi-head self-attention layers, which enhances the model's robustness against forgetting and its adaptability to new tasks. 
In contrast, while the gains from selective retraining are less dramatic, it still contributes to overall the performance enhancement, particularly by stabilizing trained parameters in specific scenarios.

\pgfplotscreateplotcyclelist{colorlist3}{%
    {blue, mark=*},
    {green, mark=*},
    {red, mark=triangle*},
}

\begin{figure}[h]
\centering
\setlength{\abovecaptionskip}{0.2cm}
\begin{tikzpicture}
\begin{axis}[
    width=0.4\textwidth,
    height=5cm,
    xlabel={task id},
    ylabel={accuracy (\%)},
    cycle list name=colorlist3,
    label style={font=\small, inner sep=0pt},
    tick label style={font=\small},
    grid=major,
    grid style={dotted, gray},
    legend pos=south west,
    legend style={fill=none, font=\fontsize{7}{10}\selectfont,legend columns=3, draw=gray!40,at={(0.04, 0.2)}},
    xlabel style={font=\small, at={(axis description cs:0.5,0.04)}},
    ylabel style={font=\small, at={(axis description cs:0.08,0.5)}},
    xmin=1, ymin=0, 
    xmax=9, ymax=100,
    xtick={1, 2, 3, 4, 5,6,7,8,9}, 
    ytick={0, 20, 40, 60, 80, 100}, 
]
\addplot coordinates {(1,100.0) (2,88.33) (3,75.78) (4,62.92) (5,59.0) (6,58.33) (7,60.71) (8,61.38) (9,48.3)};
\addlegendentry{Prefix-Z}
\addplot coordinates {(1,100.0) (2,91.67) (3,77.56) (4,64.83) (5,64.0) (6,59.22) (7,54.43) (8,57.29) (9,50.0)};
\addlegendentry{Prefix-R}
\addplot coordinates {(1,100.00) (2,96.67) (3,87.22) (4,65.42) (5,66.11) (6,60.00) (7,58.33) (8,60.54) (9,53.52)};
\addlegendentry{Prefix-A}
\end{axis}
\end{tikzpicture}
\caption{Ablation study examining the impact of parallel adapters on long tasks using the MMFi dataset.}
\label{fig_ablation_adapter}
\vspace{-0.38cm} 
\end{figure}
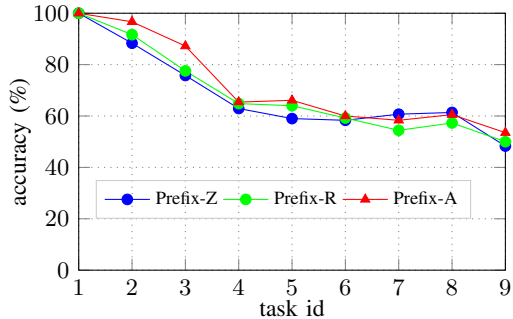

\subsubsection{Effect of Parallel Adapter}
Fig. \ref{fig_ablation_adapter} shows that the parallel adapter initialization (Prefix-A) outperforms zero (Prefix-Z) and random (Prefix-R) initializations in WECAR.
Specifically, Prefix-A achieves an average accuracy of 71.97\%, compared to 69.00\% for Prefix-R and 68.30\% for Prefix-Z.
These results underscore the effectiveness of parallel adapter initialization in enhancing prefix handling and overall the model performance.

\section{Conclusion}
We have presented WECAR, an end-edge collaborative inference and training framework for WiFi-based continuous HAR that simultaneously addresses dynamic environment adaptation and resource constraints in edge deployment.
 The proposed architecture features two key technical contributions: (1) a parameter-efficient dynamic continual learning mechanism that enables environment-aware model evolution, and (2) a dual-phase knowledge distillation strategy optimized for resource-constrained edge devices.
Implemented on heterogeneous hardware, our framework demonstrates superior performance with 10\%  average accuracy improvement across all the tasks while maintaining a forgetting rate below 21\%, outperforming state-of-the-art methods by significant margins.
What's more, WECAR achieves these advancements with only 33\% parameter count in comparison to the baseline approaches, and cutting model footprints of baseline models by 90\% in specific scenarios. 
 Extensive experiments validate its practical viability for privacy-sensitive applications, particularly demonstrating robust operation under strict computational and memory constraints. 
 The presented framework establishes new design paradigms for adaptive edge intelligence systems in ubiquitous sensing applications.

\bibliographystyle{IEEEtran}
\bibliography{IEEEabrv,IEEE}

\newpage

\end{document}